\documentclass[conference]{IEEEtran}
\usepackage{cite}
\usepackage[dvipsnames]{xcolor}
\usepackage{amsmath,amssymb,amsfonts}
\usepackage{algorithmic}
\usepackage{graphicx}
\usepackage{textcomp}
\usepackage{gensymb}
\usepackage{float}
\usepackage{booktabs}
\usepackage{subcaption}
\usepackage{sidecap}
\usepackage{copyrightbox}
\usepackage{multirow}

\DeclareMathOperator{\E}{\mathbb{E}}

\def\BibTeX{{\rm B\kern-.05em{\sc i\kern-.025em b}\kern-.08em
		T\kern-.1667em\lower.7ex\hbox{E}\kern-.125emX}}
\begin{document}
	
	\title{SAR Image Synthesis with Diffusion Models\\
	}
	
	\author{\IEEEauthorblockN{Denisa Qosja, Simon Wagner, Daniel O'Hagan}
		\IEEEauthorblockA{
			Fraunhofer Institute for High Frequency Physics and Radar Techniques FHR \\
			\{denisa.qosja, simon.wagner, daniel.ohagan\}@fhr.fraunhofer.de}

	}
	
	\IEEEoverridecommandlockouts
	\IEEEpubid{\makebox[\columnwidth]{978-1-5386-5541-2/18/\$31.00~\copyright2018 IEEE \hfill} \hspace{\columnsep}\makebox[\columnwidth]{ }}
	
	\maketitle
	
	\IEEEpubidadjcol
	
	\begin{abstract}
		
		In recent years, diffusion models (DMs) have become a popular method for generating synthetic data. By achieving samples of higher quality, they quickly became superior to generative adversarial networks (GANs) and the current state-of-the-art method in generative modeling. However, their potential has not yet been exploited in radar, where the lack of available training data is a long-standing problem. In this work, a specific type of DMs, namely denoising diffusion probabilistic model (DDPM) is adapted to the SAR domain. 
		We investigate the network choice and specific diffusion parameters for conditional and unconditional SAR image generation. 
		%
		In our experiments, we show that DDPM qualitatively and quantitatively outperforms state-of-the-art GAN-based methods for SAR image generation. Finally, we show that DDPM profits from pretraining on large-scale clutter data, generating SAR images of even higher quality.
		
	\end{abstract}
	

	\section{Introduction}
	\label{section:first}
	Synthetic Aperture Radar (SAR) technology provides high-resolution imaging data used in various applications such as surveillance, defense, remote sensing, among others. Clear images, regardless of light and weather conditions, are benefits that make SAR such a valuable sensor. 
	However, it comes with high costs of acquisition, which makes the availability of data scarse. That sets off a problem in the era of deep learning, where training neural networks requires large-scale datasets in order to appropriately extract information from the data while maintaining an adequate generalization. Target recognition with SAR~\cite{wagner2016sar, qosja2023} is therefore limited and advances at a slow pace relative to advances in comparable fields, such as LiDAR. 
	
	\begin{figure}[t!]
		\centering
		\includegraphics[width=0.47\textwidth]{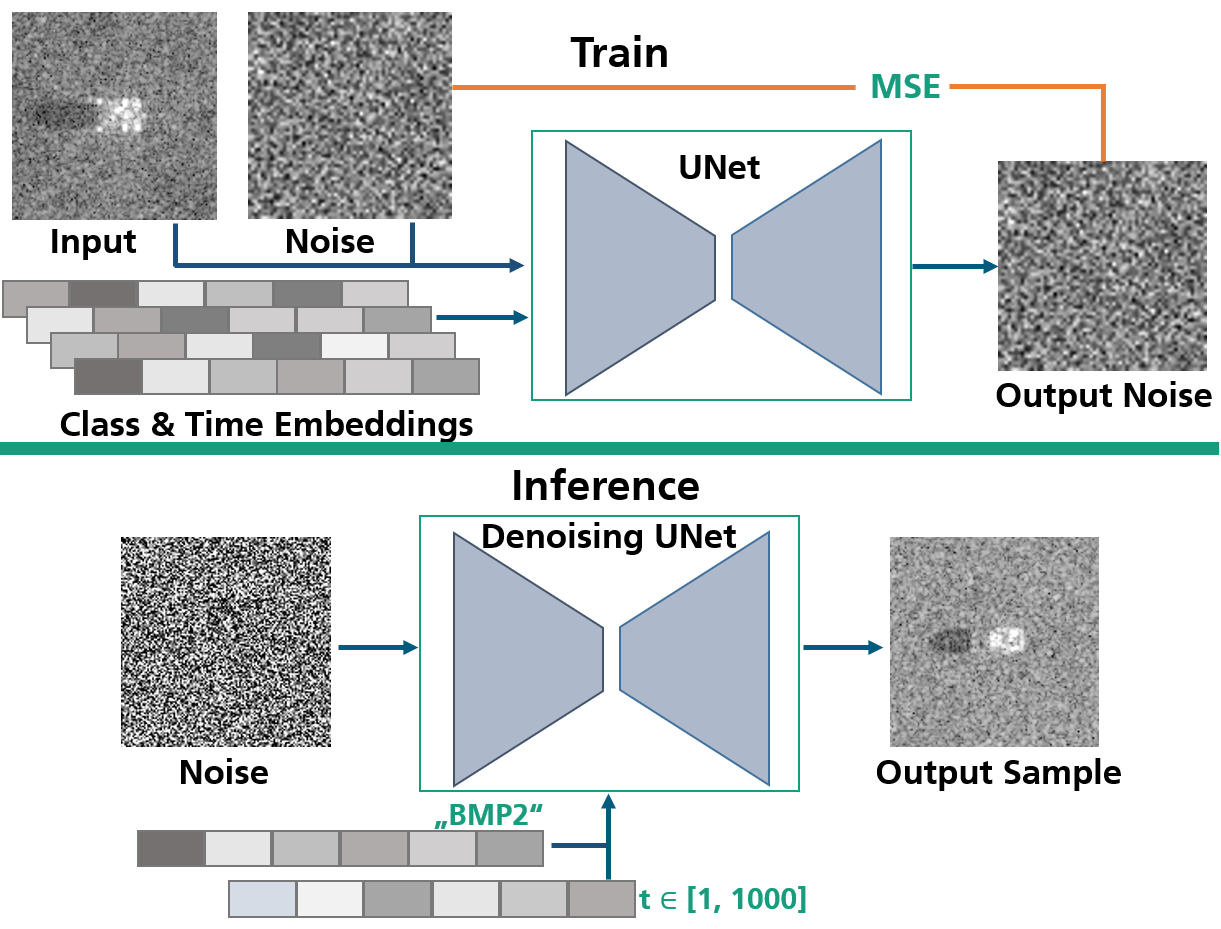}
		\vspace{0.2cm}
		\caption{Training and inference phases for DDPM. During training, input SAR data infused with noise, timestep and class embeddings are fed to the UNet model, which learns to predict the noise added to the input samples. During inference, noise is sampled from a normal distribution, fed to the UNet and  iteratively denoised to finally obtain a SAR image. }
		\label{fig:teaser_img}
		\vspace{-0.5cm}
	\end{figure}
	
	\begin{figure*}[t!]
		\centering
		\includegraphics[width=1\textwidth]{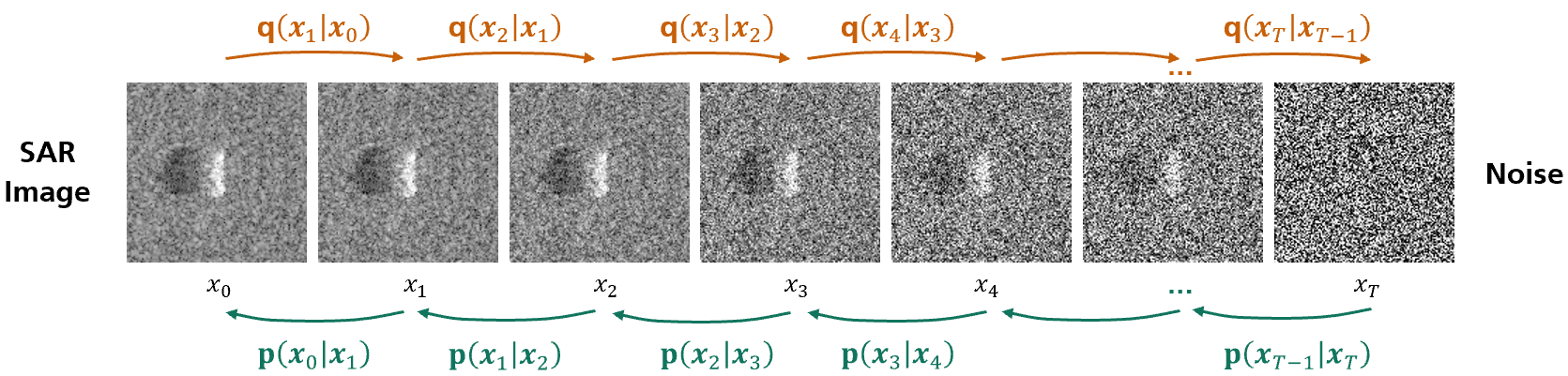}
		\caption{DDPM processing illustrated with the MSTAR dataset. The model is comprised of two processes: a \textcolor{Bittersweet}{\textbf{forward diffusion process}} that adds noise to the data and learns it for each timestep $t$; and a \textcolor{ForestGreen}{\textbf{reverse denoising process}} that draws samples from a Gaussian noise at time $T$ and iteratively denoises its input to generate new data samples.}
		\label{fig:ddpm_idea}
	\end{figure*} 
	
	To overcome the lack of abundant radar training data, efforts have been made to obtain synthetically generated data. SAR simulators have been developed through the years to simulate either images directly from real input data ~\cite{polsarprosim, auer2016raysar, hammersimulation, cochin2008mocem} or from raw signals~\cite{franceschetti1995sar, franceschetti1992saras, mori2004time, margarit2006usage, drozdowicz2021open}, from which the images are obtained after raw data processing. 
	The major benefit of simulations is the reduced acquisition costs for the data, but also the reduced cost, time and effort in obtaining annotations, overcoming the need of manual labelling. However, there are downsides to using only simulated data. The simulators are usually specified to represent targets in a specific background, e.g. sea, earth or forests, and different simulators have to be developed in order to portray different environments. Furthermore, they do not represent the full spectrum of the real-case scenario clutter, creating datasets that are far from the realism of the world. 
	
	Generative modeling with neural networks, which has shown promise to synthesize novel SAR samples, could overcome the limitations of simulators. Generative Adversarial Networks (GANs)~\cite{gan_goodfellow} have been for a long time the state-of-the-art network in the field and continue to be developed to achieve data of high realism and fidelity. In radar, they have already been used for various purposes. In~\cite{gan_guo}, an end-to-end GAN has been employed to generate 1000 samples for each class of MSTAR dataset. They make a clear analysis of why the clutter and speckle influence the network towards mode collapse and propose a normalization method to stabilize and accelerate the training process. More sophisticated versions of GANs, such as conditional GAN, WGAN, BEGAN, etc~\cite{qin2022target, luo_minority, liu_semisupervised} have been investigated and developed to generate synthetic samples. In addition, an extensive work has been made towards image translation~\cite{fuentes2019sar, zuo2021sar} in order to convert samples from SAR to the optical domain. GANs have been used to modify the simulated data by augmentations~\cite{hwang_tripleGAN} and then using it as the training set for automatic target recognition (ATR) \cite{cui_dataAugment, camus_refining}. Despite these recent advances in applying GANs to SAR data, training GANs still remains a challenging task due to training instability, low diversity, mode collapse and difficult hyperparameter optimization~\cite{saxena2021generative}. 

	In the recent years, diffusion models~\cite{diffusion_original} have emerged as a potential method in generating high quality data. They have quickly achieved state-of-the-art results in the vision field by surpassing GANs. In the radar field, however these networks have not yet found a broad application except for a recent work in the SAR domain where speckle is removed from the data~\cite{perera2023sar}. 
	
	In this paper, we investigate Denoising Diffusion Probabilistic Models (DDPM)~\cite{ddpm} to generate SAR data. An overview of our proposed approach is shown in Fig.~\ref{fig:teaser_img}. Firstly, we describe the fundamentals of DDPM and discuss different design choices and parameters to adapt them to the SAR domain. 
	We train DDPM on the MSTAR~\cite{mstar} dataset in order to generate a large number of novel samples. With the goal of extending the amount of available labelled data, we feed the network with class information in order to generate accurate images conditioned on the given category.
	Nonetheless, both class-conditional and unconditional versions of DDPM are evaluated and compared with GANs. Additionally, we show that by pretraining DDPM on a large-scale dataset of clutter images and fine-tuning it on MSTAR targets, we can generate SAR images of even higher quality.
  
	Whereas most previous works on SAR image generation seldom report quantitative metrics, in this work, we benchmark DDPM and different GAN-based models using three popular evaluation metrics, namely Inception Score (IS)~\cite{inception_score}, Fréchet Inception Distance (FID)~\cite{fid} and Kernel Inception Distance (KID)~\cite{kid}. We qualitative and quantitative show that DDPM synthesizes MSTAR targets significantly more accurately and with higher fidelity than GANs. 
	
	In section~\ref{section:second}, the DDPM will be revised, followed by a description of the method used to generate SAR data and the metrics used to evaluate them in section~\ref{section:third}. Experimental results from various networks and comparisons amongs them are described in section~\ref{section:fourth}. Finally, the paper is concluded in section~\ref{section:fifth}. 

	\section{Denoising Diffusion Probabilistic Models}
	\label{section:second}
	DDPM was introduced by Ho et al.~\cite{ddpm}. DDPM is composed of a Markov chain, known as a forward diffusion process, where the data is infused gradually with Gaussian noise. It is followed by a reverse denoising process, with the learned noise being removed to generate a new data sample which has the same distribution as the original data. The simple idea of DDPM is displayed in Fig.~\ref{fig:ddpm_idea}. 
	
	\subsection{Forward Diffusion Process}
	
	Suppose that a training sample $\mathbf x_{0}$ is of certain distribution, denoted as $q(\mathbf x_0)$. In the forward diffusion process, Gaussian noise with variance $\beta_t \in (0, 1)$ is added gradually to the sample $\mathbf x_0$ for $T$ steps. The number of timesteps $T$ is chosen to be large enough so that a latent sample $\mathbf x_T$ with isotropic Gaussian distribution is obtained: $\mathbf x_T \sim \mathcal{N}(0, \mathbf I)$. 
	
	The posterior distribution, which transitions into latent space to provide samples injected with Gaussian noise $\mathbf x_1$, ..., $\mathbf x_T$, is defined as: 
	\begin{equation}
	q(\mathbf x_{1:T}|\mathbf x_0) = \prod_{t = 1}^{T} 	q(\mathbf x_t|\mathbf x_{t-1}),
	\end{equation}
	\begin{equation}
	q(\mathbf x_t|\mathbf x_{t-1}) := \mathcal{N}(\mathbf x_t, \sqrt{1 - \beta_t} \mathbf{x}_{t-1}, \beta_t \mathbf{I}). 
	\end{equation}
	
	To make the forward process faster, the samples $\mathbf{x}_t$ can be sampled arbitrarily at any timestep $t$. Using the notations $\alpha_t := 1 - \beta_t$ and $\bar{\alpha}_t := \prod_{s = 1}^{t} \alpha_s$, the posterior probability can be rewritten as
	\begin{equation}
	q(\mathbf x_t|\mathbf x_0) = \mathcal{N}(\mathbf x_t, \sqrt{\alpha_t} \mathbf{x}_{0}, (1 - \bar{\alpha}_t) \mathbf{I}).
	\end{equation}
	
	The posterior $q$ has no learnable parameters which makes the forward process fixed (the variances $\beta_t = 1 - \bar{\alpha_t}$ are not learnable). Therefore, the sample $\mathbf x_t$ can be defined directly as
	\begin{equation}
	\mathbf x_t = \sqrt{\bar{\alpha}_t} \mathbf x_0 + \sqrt{1 - \bar{\alpha}_t} \boldsymbol{\epsilon}, \hspace{5mm} \boldsymbol{\epsilon} \sim \mathcal{N}(0, \mathbf I). 
	\label{eq:derive_xt}
	\end{equation}

	\subsection{Reverse Denoising Process}
	
	The reverse denoising process aims to sample reversely from $\mathbf x_{T}$ through transition probabilities $q(\mathbf x_{t-1}|\mathbf x_t) $ for timesteps $T-1$ through $1$ to obtain a sample drawn from $q(\mathbf x)$. The transition $q(\mathbf x_{t-1}|\mathbf x_t) $ is a Gaussian distribution, tractable when conditioned on $\mathbf {x}_0$, defined as:
	\begin{equation}
	q(\mathbf x_{t-1} | \mathbf x_t, \mathbf x_0) = \mathcal{N}(\mathbf x_{t-1}, \tilde{\boldsymbol{\mu}}_t(\mathbf x_t, \mathbf x_0), \tilde{\beta}_t \mathbf I).
	\end{equation}
	As shown in~\cite{weng2021diffusion}, using Bayes' rule, the mean $\tilde{\boldsymbol{\mu}}_t$ and variance $\tilde{\beta}_t$ are calculated to be:
	\begin{equation}
	\sigma_t^2 = \tilde{\beta}_t = \frac{1 - \bar{\alpha}_{t-1}}{1 - \bar{\alpha}_t} \beta_t,
	\end{equation} 
	\begin{equation}
	\tilde{\boldsymbol{\mu}}_t(\mathbf x_t, \mathbf x_0) := \frac{\sqrt{\bar{\alpha}_{t-1}}\beta_t}{1-\bar{\alpha}_t} \mathbf x_0 + \frac{\sqrt{\alpha_t}(1 - \bar{\alpha}_{t-1})}{1-\bar{\alpha}_t} \mathbf x_t.
	\end{equation}

	Replacing $\mathbf x_0$ from (\ref{eq:derive_xt}):
	\begin{equation} 
	\tilde{\boldsymbol{\mu}}_t(\mathbf x_t, \mathbf x_0) = \frac{1}{\sqrt{\alpha_t}} \left(\mathbf x_t - \frac{1-\alpha_t}{\sqrt{1 - \bar{\alpha}_t}} \boldsymbol \epsilon_t \right).
	\end{equation}
	Given that the estimation of the reverse transition probability $p_{\theta}(\mathbf x_{t-1} | \mathbf x_t)$ relies on the entire data distribution, it can be approximated through a neural network: 
	\begin{equation}
	p_{\theta}(\mathbf x_{t-1} | \mathbf x_t) = \mathcal{N}(\mathbf x_{t-1}, \boldsymbol{ \mu}_\theta(\mathbf x_t, t), \boldsymbol{\Sigma}_\theta(\mathbf x_t, t)),
	\end{equation}
	where, following~\cite{ddpm}, $\boldsymbol{\Sigma}_\theta(\mathbf x_t, t) = \tilde{\beta}_t \mathbf I$ is a fixed variance, and the mean $\boldsymbol{ \mu}_\theta(\mathbf x_t, t)$ depends on a noise sample $\boldsymbol \epsilon_\theta(\mathbf x_t, t)$ which is learned by a neural network, conditioned on the current sample $\mathbf x_t$ and timestep $t$. The learning process is giuded by the objective function 
	
	\begin{equation}
	L = \E_{t, x_0, \epsilon} \left[ ||\boldsymbol \epsilon - \boldsymbol \epsilon_\theta(\mathbf x_t, t)||^2 \right] ,
	\end{equation}
	
	while the output sample is obtained as follows: 

	\begin{equation}
		\begin{aligned}
			\mathbf x_{t-1} &= \boldsymbol{ \mu}_\theta(\mathbf x_t, t) + \sigma_t \mathbf{z}, \\
			 &= \frac{1}{\sqrt{\alpha_t}} \left(\mathbf x_t - \frac{1-\alpha_t}{\sqrt{1 - \bar{\alpha}_t}} \boldsymbol \epsilon_\theta(\mathbf x_t, t) \right) + \sigma_t \mathbf{z}.
		\end{aligned}
	\end{equation}
	
	\begin{equation}
		\begin{cases}
			\mathbf{z} \sim \mathcal{N}(0, \mathbf I),& \text{if } t > 1\\
			\mathbf{z} = 0,              & \text{otherwise}.
		\end{cases}
	\end{equation}

	\subsection{Design Choices}
	
	To decide on the design of the network, we rely mostly on the original work by Ho et al.~\cite{ddpm}. 
	
	\textbf{Diffusion Steps:} $T$ is chosen to be sufficiently large, namely $T=1000$ in order to obtain a sample $\mathbf x_T$ with an isotropic Gaussian distribution. 
	
\begin{figure}[t!]
		\centering
		\includegraphics[width=0.45\textwidth]{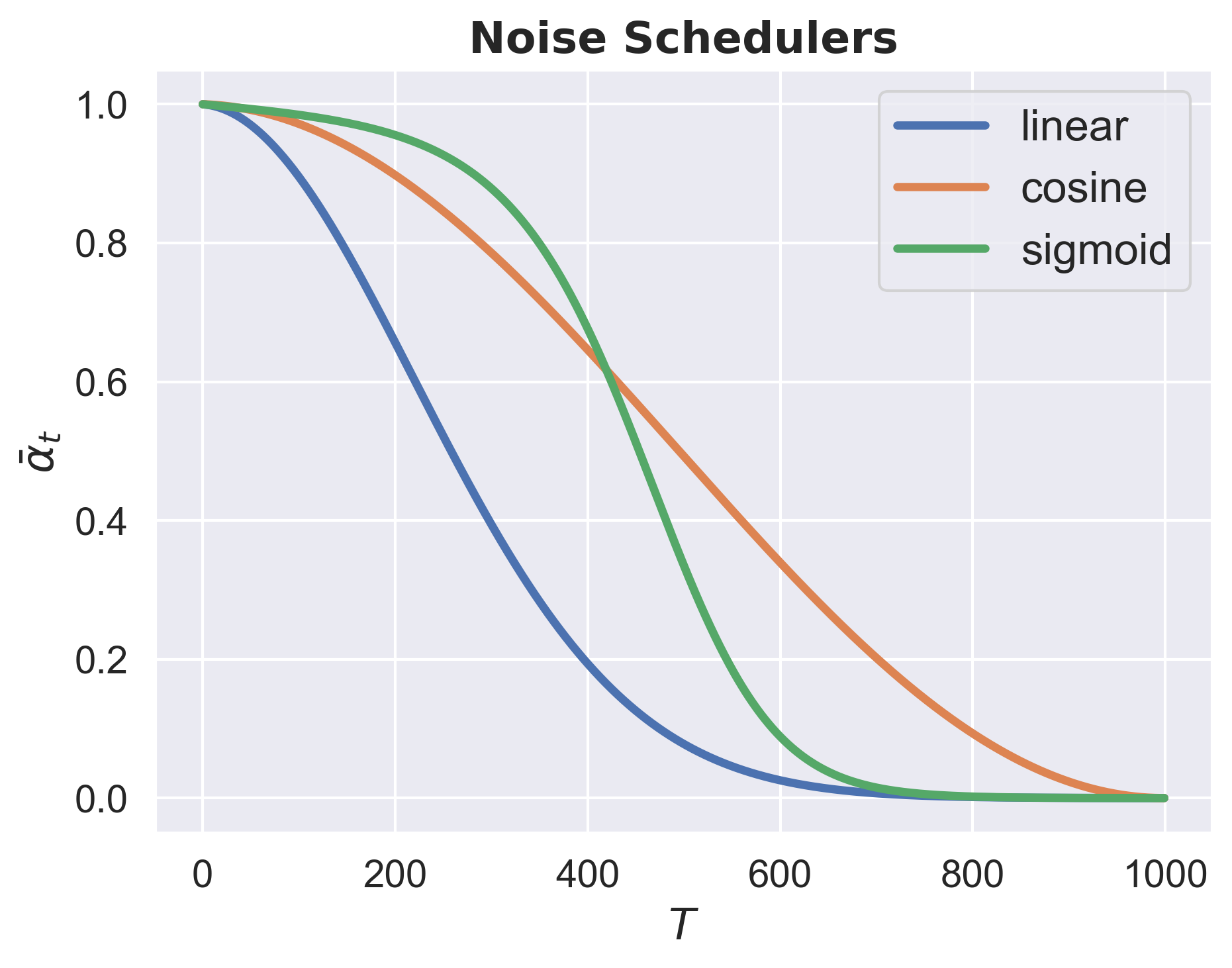}
		\caption{The behaviour of $\bar \alpha_t$ for different noise schedulers.}
		\label{fig:noise_schedulers}
	\end{figure}

	\textbf{Noise Schedule:} The noise is added to the input data in a linear manner, however it is argued in~\cite{nichol2021improved} that it does not contribute much towards the end of the diffusion process. In this work, we explore two other schedulers, namely cosine and sigmoid, with variances  as shown in Fig.~\ref{fig:noise_schedulers}. Based on our experiments, the linear manner of infusing noise performs the best in SAR. We argue that the quick decay of the noise levels occuring in early time steps in the linear scheduler leads to an improved performance.
	
	$\boldsymbol{\beta}$ \textbf{Range:} The variance range is decided to be in between $\beta_1 = 10^{-4}$ and $\beta_T = 0.02$, considering that data has to be scaled to $[-1, 1]$ before being fed to the network.
	
	\textbf{Conditional DDPM:} 
	Conditional generative models are designed to generate data based on condition presented to them. Since targets from certain classes are desired, the condition is simply the class information embedded into each downsampling and upsampling block of the network. 
	
	\subsection{Model - UNet}
	
	The model used to learn the Gaussian noise added to the data samples is a UNet~\cite{unet} composed of eight residual blocks in the compression and in the extension side, with three downsampling and upsampling layers respectively. The network is equiped with residual connections to fight vanishing gradients and skip connections to recover high detailed information at the extention part.  
	
	Following~\cite{ddpm}, the network includes group normalization layers~\cite{group_normalization}, self-attention blocks at $32 \times 32$ feature map resolution, a dropout layer with $p = 0.3$, time and label embeddings, and Swish activation function~\cite{swish}.

	\section{SAR Data Generation}
	\label{section:third}
	In this work, we make use of MSTAR dataset which is composed of a training set with 2693 images and a test set with 4357 images. We aim to extend the training set by generating 10k synthetic samples. Before being fed to the network, the data should be normalized to range $[-1, 1]$. 
	
	\subsection{MSTAR Clutter}
	Since the scale of the MSTAR dataset is small, we propose to exploit the larger amount of unlabelled clutter images for pretraining generative models. 

\begin{figure}[t!]
		
		\centering
		\includegraphics[width=.16\textwidth]{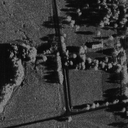}\hfill
		\includegraphics[width=.16\textwidth]{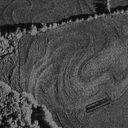}\hfill
		\includegraphics[width=.16\textwidth]{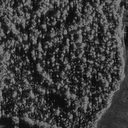}
		
		\caption{The clutter data depicts the surroundings of MSTAR targets, and illustrates scenarios such as fields, trees and roads. }
		\label{fig:clutter}
		
	\end{figure}

	We employ the MSTAR clutter data, which includes large scenarios of the environment where MSTAR targets were acquired. Scenes like trees, fields and roads are depicted in 100 images of large sizes, mostly $1784\times1476$. These images are preprocessed into same shapes as MSTAR targets of $128\times128$, and converted into logarithmic scale, resulting in 14.3k training samples. Examples of unprocessed clutter data are shown in Fig.~\ref{fig:clutter}. 
	
	With the goal of generating higher quality images, we propose to first pretrain generative models on the MSTAR clutter dataset, and then fine-tune them on MSTAR targets for class-conditional image generation. 
	
	\subsection{GANs Architectures}
	
	We compare DDPM with three popular GAN variants which have shown success for synthesising MSTAR images; namely Deep Convolutional GAN (DCGAN)~\cite{dcgan}, Self-Attention GAN (SAGAN)~\cite{sagan} and Wasserstein GAN (WGAN)~\cite{wgan}.  
	
	DCGAN is the first version of GANs to introduce strided convolutional layers in the architecture. It enforced the usage of batch normalization stabilizing the training, and proposed leaky ReLU as the activation function at the discriminator while maintaining ReLU at the generator. 
	
	SAGAN introduced a self-attention mechanism in convolutional GANs to model long-range dependencies across image regions. In addition, it comes with some different design choices, such as spectral normalization~\cite{spectralnorm}, class conditioning inside the batch normalization layer at the generator, and the hinge version of the adversarial loss. Other changes include using different learning rates for the generator and discriminator. 
	
	WGAN was introduced as a version of GANs which uses the Wasserstein's distance as the loss function and improves the mode collapse issue. In this work, we combine the SAGAN architecture with the Wasserstein's loss. In addition, we use gradient penalty instead of weight clipping. 
	
	\subsection{Evaluation Protocol}
	
	When measuring the performance of a generative model, two characteristics of the synthetic data should be considered: fidelity and diversity. It is important that the fake samples are of high quality and as close as possible to the real data, and that the generated set is of high diversity, covering the distribution of the real set.  
	
	Since a standardized evaluation protocol has not yet been established in SAR image generation, we define the following evaluation pipeline. 
	Given a generative model (i.e. DDPM or GAN) trained on MSTAR, we generate 10k novel samples, and evaluate them using the IS, FID and KID metrics, which employ an InceptionNetV3~\cite{inception_net} pretrained on MSTAR.

	IS uses a pretrained InceptionNetV3 to predict conditional probabilities of the generated samples. A high classification value implies a sample with high quality. The diversity is measured as the Kullback-Leibler (KL) divergence between the marginal and conditional probability distributions. Both outcomes are combined into scores that show good results from the generative network. A large combination of both outcomes indicates generated samples of high quality and diversity. However, the IS does not consider the statistics of the real data and also does not measure the intra-class diversity. 

	FID was developed to improve on the drawbacks of the IS. The InceptionNetV3 extracts features of dimensionality $2048$ from both real and fake data, which are then embedded into multivariate Gaussian distributions. Fréchet distance outputs a similarity score in the distribution between measured data and generated one. The smaller the score, the closer the distribution is. However, the statistics to calculate the distance (mean and variance) make the FID dependent on the data size. 
	
	KID overcomes the data size issue, by calulating the maximum mean discrepancy between features of the real and generated data. 
	
	\setlength{\tabcolsep}{8pt}
	\begin{table}
		\centering
		\caption{Evaluation of unconditional (top) and conditional (bottom) GANs and DDPM. DDPM clearly outperforms GAN-based models.}
		\label{table:comparison_gan_dm}
		\vspace{0.02cm}
		\small
		\begin{tabular}{lccc}
			\toprule
			\textbf{Model}      	&  \textbf{IS} $\uparrow$  &  \textbf{FID} $\downarrow$ &  \textbf{KID} $\downarrow$	\\
			\midrule
			DCGAN\cite{dcgan}    &	4.19 & 7.10 & 0.0053 \\
			SAGAN\cite{sagan} 	 &  7.00    & 3.11    & 0.0025 \\
			DDPM\cite{ddpm}     &\textbf{8.64} & \textbf{0.77} & \textbf{0.0004} \\
			\midrule
			\midrule
			cDCGAN \cite{dcgan} 		& 7.60        & 3.60   &	0.0035  	\\
			cSAGAN\cite{sagan} 		& 9.39        & 2.46			& 0.0017 	\\
			cWGAN\cite{wgan} 		& 7.00		  & 1.59			& 0.0008	\\
			cDDPM~\cite{ddpm}	& \textbf{9.77}   & \textbf{0.38}	& \textbf{0.0003} \\
			\bottomrule
		\end{tabular}		
	\end{table}

	\section{Experiments}
	\label{section:fourth}
	 In this section, we evaluate the performance of DDPM and compare it to the GAN variants. All networks are trained both for conditional and unconditional SAR image generation for 200 epochs, using a batch size of 32. DDPM has an initial channel size 64 which doubles at each downsampling step. For all GANs, this initial value is set to 16. 
 	 The pretraining on clutter data is set for 500 epochs, followed by 200 epochs of fine-tuning on MSTAR for class-conditional generation. These parameters are chosen through empirical validation. 
	 
\begin{figure}[t!]
		\begin{subfigure}{0.5\textwidth}
			\includegraphics[width=.2\textwidth]{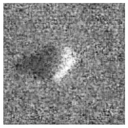}\hfill
			\includegraphics[width=.2\textwidth]{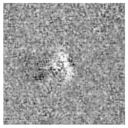}\hfill
			\includegraphics[width=.2\textwidth]{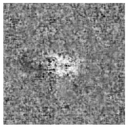}\hfill
			\includegraphics[width=.2\textwidth]{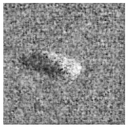}\hfill
			\includegraphics[width=.2\textwidth]{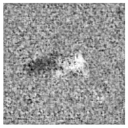}
			\caption{cDCGAN} \label{fig:dcgan}
		\end{subfigure}%
		
		\begin{subfigure}{0.5\textwidth}
			\includegraphics[width=.2\textwidth]{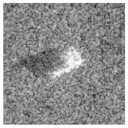}\hfill
			\includegraphics[width=.2\textwidth]{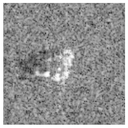}\hfill
			\includegraphics[width=.2\textwidth]{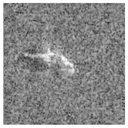}\hfill
			\includegraphics[width=.2\textwidth]{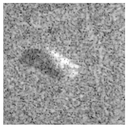}\hfill
			\includegraphics[width=.2\textwidth]{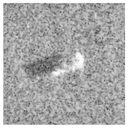}
			\caption{cSAGAN} \label{fig:sagan}
		\end{subfigure}%
		
		\begin{subfigure}{0.5\textwidth}
			\includegraphics[width=.2\textwidth]{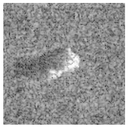}\hfill
			\includegraphics[width=.2\textwidth]{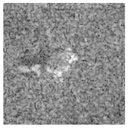}\hfill
			\includegraphics[width=.2\textwidth]{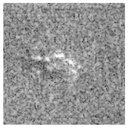}\hfill
			\includegraphics[width=.2\textwidth]{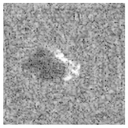}\hfill
			\includegraphics[width=.2\textwidth]{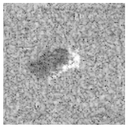}
			\caption{cWGAN} \label{fig:wgan}
		\end{subfigure}%
		
		\begin{subfigure}{0.5\textwidth}
			\includegraphics[width=.2\textwidth]{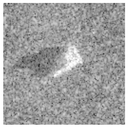}\hfill
			\includegraphics[width=.2\textwidth]{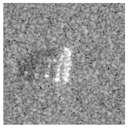}\hfill
			\includegraphics[width=.2\textwidth]{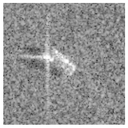}\hfill
			\includegraphics[width=.2\textwidth]{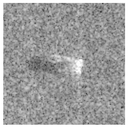}\hfill
			\includegraphics[width=.2\textwidth]{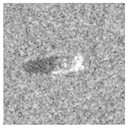}
			\caption{cDDPM} \label{fig:ddpm}
		\end{subfigure}%
		\caption{Images generated by class-conditioned generative networks. The images generated by cDDPM represent the background in a more accurate way and reconstruct the targets more precisely.}
		\label{fig:class_conditioned_results}
	\end{figure}

	 In Fig.~\ref{fig:class_conditioned_results}, a few samples from all class-conditioned architectures are shown. The images in each row belong to the classes M548, M35 (x2), 2S1 and BTR70 from left to right. The synthetic data obtained by DCGAN includes visible artifacts, unclear target shapes and ring-like patterns around the targets. The quality in SAGAN-generated data is improved, but similarly to WGAN, they still lack high fidelity. Most of the artifacts present in GAN-generated samples are resolved by DDPM, which can also represent more accurately reflections.  
	 
	 Quantitative results in terms of IS, FID and KID are given in Table~\ref{table:comparison_gan_dm}, where best achieved results are in bold. On the top part, the values are obtained from the unconditional architectures, while on the bottom from their conditional counterpart. As expected, the conditional versions improve the quality and diversity of the data with respect to the unconditional ones. In both cases, DDPM achieves the best performance with a significant improvement with respect to the GAN architectures.
	 
\begin{figure}[t!]
		\centering
		\begin{subfigure}{0.095\textwidth}
			\includegraphics[width=\textwidth]{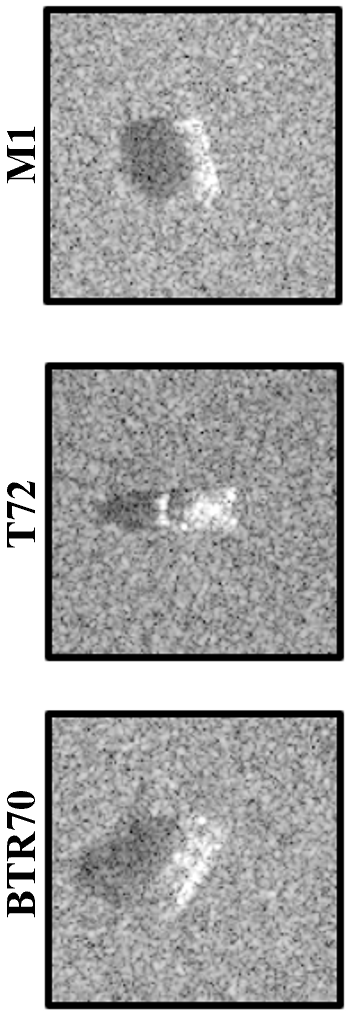}
			\caption{GT} \label{fig:ddpm_ground truth}
			\label{subfig:GT}
		\end{subfigure}%
		\hspace*{0.25mm}
		\begin{subfigure}{0.405\textwidth}
			\includegraphics[width=\textwidth]{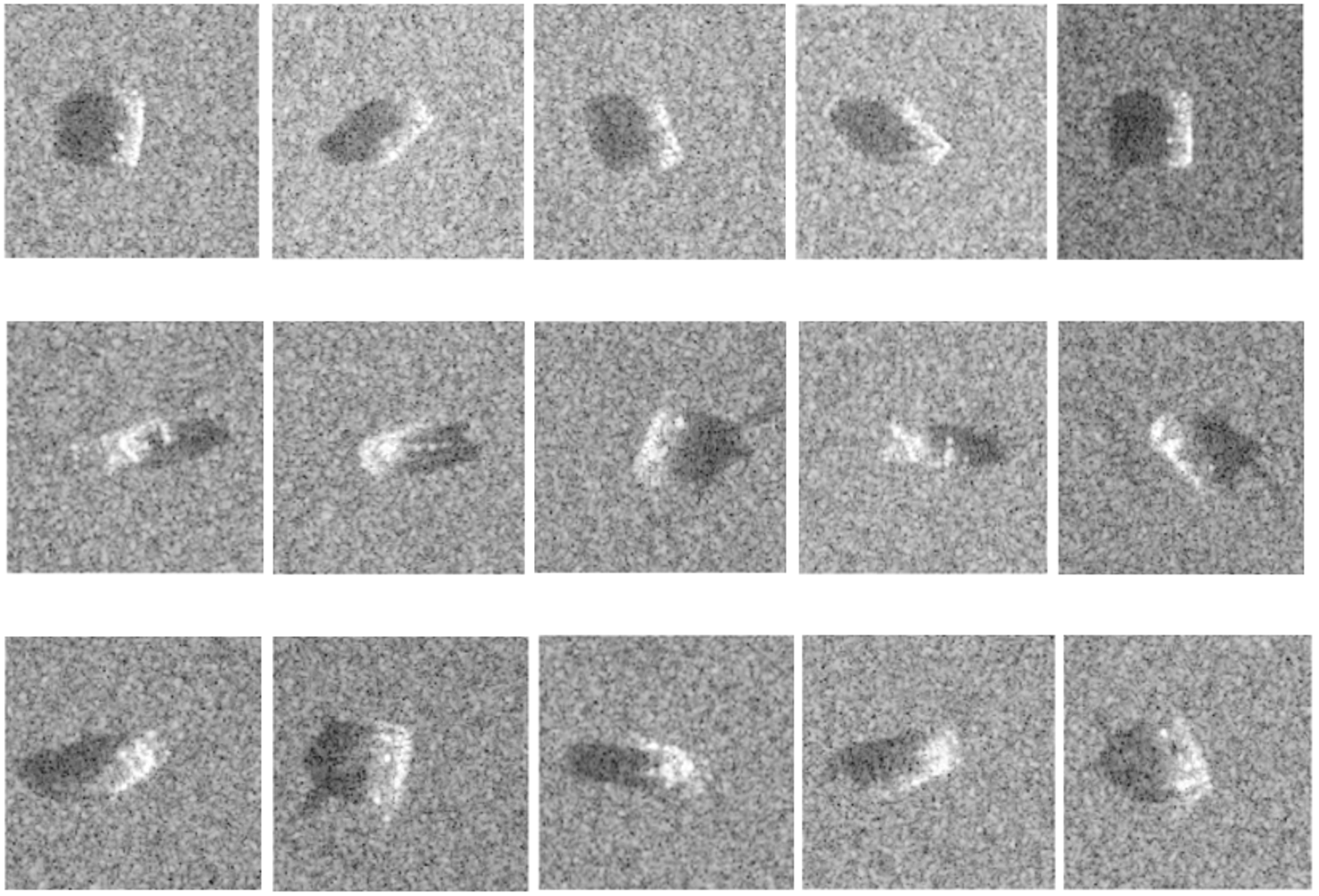}
		\caption{Generated Targets} \label{fig:ddpm_generated_targets_}
		\end{subfigure}%
		\vspace{-0.1cm}
		\caption{\textbf{(a)} Ground truth MSTAR images of classes M1, T72 and BTR70. \textbf{(b)} Synthetic samples generated by DDPM when conditioned on the corresponding class.}
		\label{fig:ddpm_images}
	\end{figure}

	 For a closer look on the performance of DDPM in SAR, in Fig.~\ref{fig:ddpm_images}, generated MSTAR images of categories M1, T72 and BTR70 are shown alongside ground truth samples from the corresponding class. We show that DDPM is able to synthesize high diversity data from various azimuth angles. DDPM generates high quality images with target shapes reconstructed accurately and background represented adequately.
	
	\begin{table}
		\centering
		\caption{Evaluation of pretrained GANs and DDPM.}
		\label{table:comparison_pretrained_gan_dm}
		\vspace{0.02cm}
		\small
		\begin{tabular}{lcccc}
			\toprule
			& \textbf{Model} & \textbf{IS}  $\uparrow$ & \textbf{FID} $\downarrow$ & \textbf{KID} $\downarrow$ \\ 
			\cmidrule[0.7pt] (l{8em}r{0em}){1-5}
			\multirow{2}{*}{cDCGAN\cite{dcgan}} & no pretraining & \textbf{7.60}    & \textbf{3.60}   &	\textbf{0.0035}  	\\ 
			& pretraining  & 3.87 & 9.93 & 0.0060 \\
			\midrule
			\multirow{2}{*}{cDDPM ~\cite{ddpm}} & no pretraining	& \textbf{9.77}   & 0.38	& 0.0003     \\
			& pretraining & 9.74  & \textbf{0.23} & \textbf{0.0001} \\
			\bottomrule
			\vspace*{-0.5cm}
		\end{tabular}		
	\end{table}

\begin{figure}[t!]
		\vspace{-0.2cm}
		\begin{subfigure}{0.5\textwidth}
			\includegraphics[width=.25\textwidth]{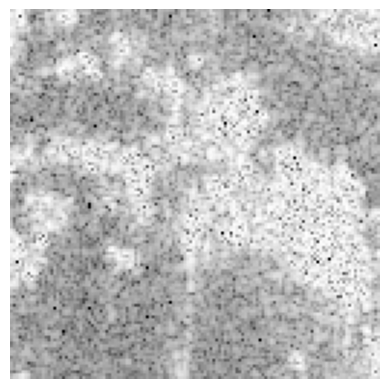}\hfill
			\includegraphics[width=.25\textwidth]{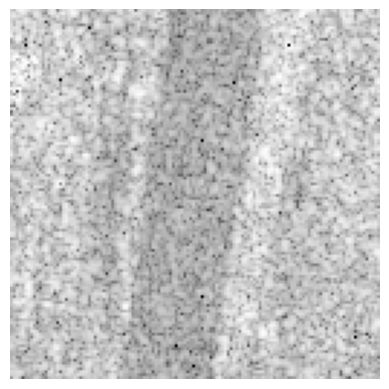}\hfill
			\includegraphics[width=.25\textwidth]{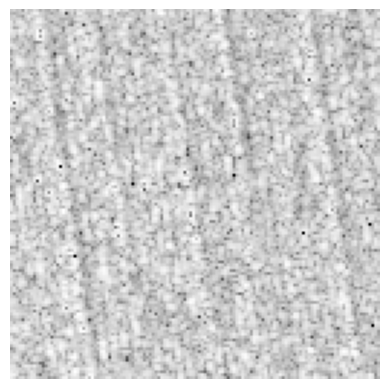}\hfill
			\includegraphics[width=.25\textwidth]{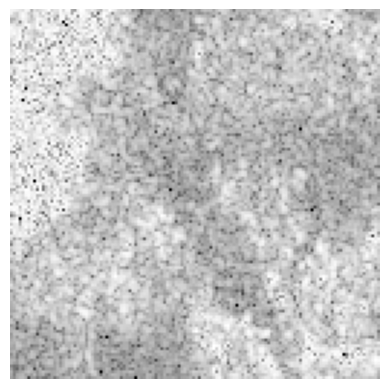}
			\caption{Generated Clutter} \label{fig:generated_clutter}
		\end{subfigure}%
		
		\begin{subfigure}{0.5\textwidth}
			\includegraphics[width=.25\textwidth]{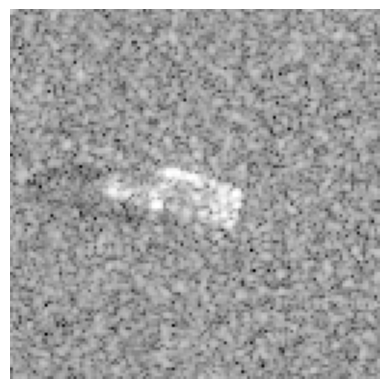}\hfill
			\includegraphics[width=.25\textwidth]{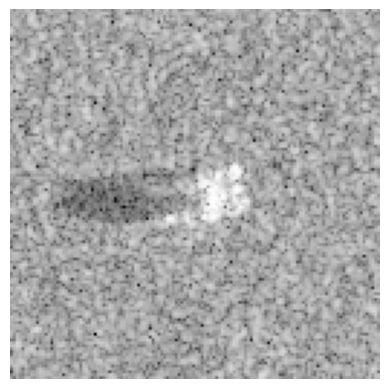}\hfill
			\includegraphics[width=.25\textwidth]{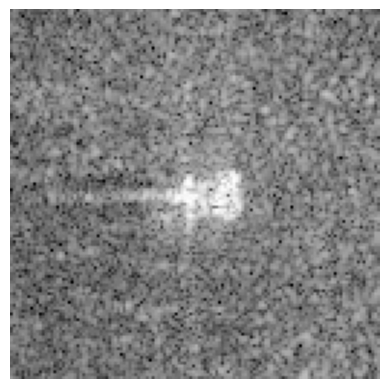}\hfill
			\includegraphics[width=.25\textwidth]{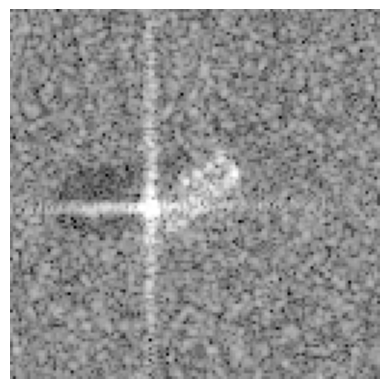}
			\caption{cDDPM} \label{fig:unpretrained_ddpm}
		\end{subfigure}%
		
		\begin{subfigure}{0.5\textwidth}
			\includegraphics[width=.25\textwidth]{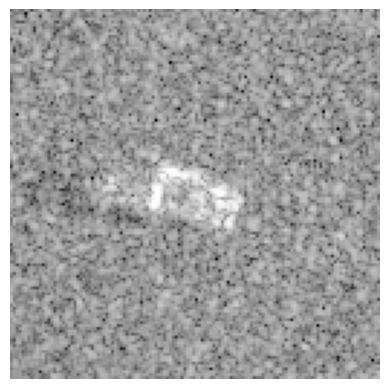}\hfill
			\includegraphics[width=.25\textwidth]{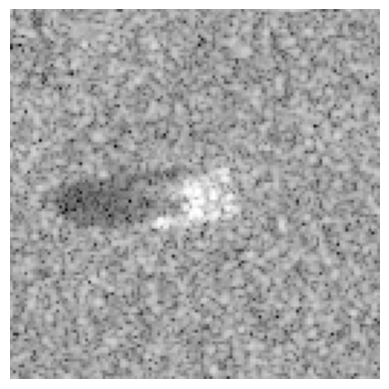}\hfill
			\includegraphics[width=.25\textwidth]{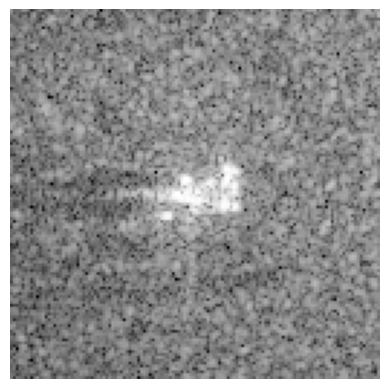}\hfill
			\includegraphics[width=.25\textwidth]{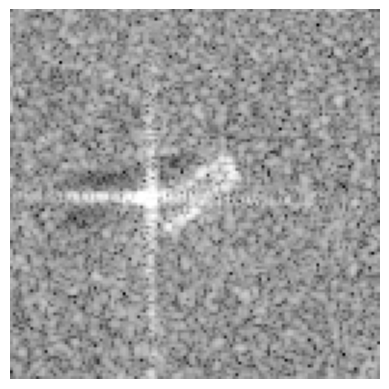}
			\caption{Pretrained cDDPM} \label{fig:pretrained_ddpm}
		\end{subfigure}%
		\vspace{-0.1cm}
		\caption{\textbf{(a)} Clutter generated during pretraining by DDPM. \textbf{(b)} Targets generated by unpretrained conditional DDPM. \textbf{(c)}~Targets generated by DDPM pretrained on clutter and fine-tuned on MSTAR for conditional image generation.}
		\label{fig:pretrained_results}
		\vspace{-0.3cm}
	\end{figure}

	In Table~\ref{table:comparison_pretrained_gan_dm}, we compare DCGAN and DDPM models first pretrained with clutter background and then fine-tuned on MSTAR, with their unpretrained counterparts. 
	Whereas clutter pretraining deteriorates the results for DCGAN, DDPM highly benefits from the pretraining, showing superior quantitative performance. 
	Fig.~\ref{fig:generated_clutter} depicts images of clutter generated by DDPM during pretraining. We can clearly distinguish different clutter scenarios, such as trees, fields and roads. 
	Fig~\ref{fig:unpretrained_ddpm} and \ref{fig:pretrained_ddpm} show images generated by the unpretrained and pretrained DDPM, respectively, using the same class condition. Despite the superior quantitative performance achieved by pretraining, both models generate realistic SAR samples.

	\section{Conclusions}
	\label{section:fifth}
	In this work, the diffusion models are brought to the SAR domain for the purpose of image synthesis. We showed that DDPM quantitatively outperforms GAN-based models on the MSTAR dataset on three evaluation metrics, while generating samples of higher quality. 
	Additionally, we showed that DDPM can leverage large amounts of unlabelled clutter data for pretraining, leading to higher quality results.
	As future work, we aim to use our high quality synthetic data to extend the small-scale available datasets for downstream tasks, such as target classification and detection. Furthermore, we plan to explore DDPM for other imaging tasks, such as image translation and super-resolution. 
	
	\bibliographystyle{IEEEtran}
	\bibliography{bibliography}

\begin{thebibliography}{10}
\providecommand{\url}[1]{#1}
\csname url@samestyle\endcsname
\providecommand{\newblock}{\relax}
\providecommand{\bibinfo}[2]{#2}
\providecommand{\BIBentrySTDinterwordspacing}{\spaceskip=0pt\relax}
\providecommand{\BIBentryALTinterwordstretchfactor}{4}
\providecommand{\BIBentryALTinterwordspacing}{\spaceskip=\fontdimen2\font plus
\BIBentryALTinterwordstretchfactor\fontdimen3\font minus
  \fontdimen4\font\relax}
\providecommand{\BIBforeignlanguage}[2]{{%
\expandafter\ifx\csname l@#1\endcsname\relax
\typeout{** WARNING: IEEEtran.bst: No hyphenation pattern has been}%
\typeout{** loaded for the language `#1'. Using the pattern for}%
\typeout{** the default language instead.}%
\else
\language=\csname l@#1\endcsname
\fi
#2}}
\providecommand{\BIBdecl}{\relax}
\BIBdecl

\bibitem{wagner2016sar}
S.~A. Wagner, ``{SAR ATR by a Combination of Convolutional Neural Network and
  Support Vector Machines},'' \emph{IEEE Transactions on Aerospace and
  Electronic Systems}, 2016.

\bibitem{qosja2023}
D.~Qosja, S.~Wagner, and S.~Br{\"u}ggenwirth, ``{Benchmarking Convolutional
  Neural Network Backbones for Target Classification in SAR},'' in \emph{2023
  IEEE Radar Conference (RadarConf23)}, 2023.

\bibitem{polsarprosim}
M.~L. Williams, ``{PolSARproSim - A Coherent Polimetric SAR Simulation of
  Forests for PolSARproSim},'' \emph{European Space Agency}, 2006.

\bibitem{auer2016raysar}
S.~Auer, R.~Bamler, and P.~Reinartz, ``{RaySAR-3D SAR Simulator: Now Open
  Source},'' in \emph{IEEE International Geoscience and Remote Sensing
  Symposium (IGARSS)}.\hskip 1em plus 0.5em minus 0.4em\relax IEEE, 2016, pp.
  6730--6733.

\bibitem{hammersimulation}
H.~Hammer and K.~Schulz, ``{Coherent Simulation of SAR Images}.''\hskip 1em
  plus 0.5em minus 0.4em\relax International Society for Optics and Photonics,
  2009.

\bibitem{cochin2008mocem}
C.~Cochin, P.~Pouliguen, B.~Delahaye, D.~le~Hellard, P.~Gosselin, and
  F.~Aubineau, ``{MOCEM-An'all in one'tool to simulate SAR image},'' in
  \emph{7th European Conference on Synthetic Aperture Radar}, 2008.

\bibitem{franceschetti1995sar}
G.~Franceschetti, M.~Migliaccio, and D.~Riccio, ``{The SAR Simulation: An
  Overview},'' in \emph{International Geoscience and Remote Sensing Symposium},
  1995.

\bibitem{franceschetti1992saras}
G.~Franceschetti, M.~Migliaccio, D.~Riccio, and G.~Schirinzi, ``{SARAS: A
  Synthetic Aperture Radar(SAR) Raw Signal Simulator},'' \emph{IEEE
  Transactions on Geoscience and Remote Sensing}, 1992.

\bibitem{mori2004time}
A.~Mori and F.~De~Vita, ``{A Time-Domain Raw signal Simulator for
  Interferometric SAR},'' \emph{IEEE Transactions on Geoscience and Remote
  Sensing}, 2004.

\bibitem{margarit2006usage}
G.~Margarit, J.~J. Mallorqui, J.~M. Rius, and J.~Sanz-Marcos, ``{On the Usage
  of GRECOSAR, an Orbital Polarimetric SAR Simulator of Complex Targets, to
  Vessel Classification Studies},'' \emph{IEEE Transactions on Geoscience and
  Remote Sensing}, 2006.

\bibitem{drozdowicz2021open}
J.~Drozdowicz, ``{The Open-Source Framework for 3D Synthetic Aperture Radar
  Simulation},'' \emph{IEEE Access}, 2021.

\bibitem{gan_goodfellow}
I.~Goodfellow, J.~Pouget-Abadie, M.~Mirza, B.~Xu, D.~Warde-Farley, S.~Ozair,
  A.~Courville, and Y.~Bengio, ``{Generative Adversarial Nets},''
  \emph{Advances in neural information processing systems}, 2014.

\bibitem{gan_guo}
J.~Guo, B.~Lei, C.~Ding, and Y.~Zhang, ``{Synthetic Aperture Radar Image
  Synthesis by using Generative Adversarial Nets},'' \emph{IEEE Geoscience and
  Remote Sensing Letters}, 2017.

\bibitem{qin2022target}
J.~Qin, Z.~Liu, L.~Ran, R.~Xie, J.~Tang, and Z.~Guo, ``{A Target SAR Image
  Expansion Method Based on Conditional Wasserstein Deep Convolutional GAN for
  Automatic Target Recognition},'' \emph{IEEE Journal of Selected Topics in
  Applied Earth Observations and Remote Sensing}, 2022.

\bibitem{luo_minority}
Z.~Luo, X.~Jiang, and X.~Liu, ``{Synthetic Minority Class Data by Generative
  Adversarial Network for Imbalanced SAR Target Recognition},'' in \emph{IEEE
  International Geoscience and Remote Sensing Symposium IGARSS}.\hskip 1em plus
  0.5em minus 0.4em\relax IEEE, 2020.

\bibitem{liu_semisupervised}
X.~Liu, Y.~Huang, C.~Wang, J.~Pei, W.~Huo, Y.~Zhang, and J.~Yang,
  ``{Semi-Supervised SAR ATR via Conditional Generative Adversarial Network
  with Multi-Discriminator},'' in \emph{{IEEE International Geoscience and
  Remote Sensing Symposium IGARSS}}, 2021.

\bibitem{fuentes2019sar}
M.~Fuentes~Reyes, S.~Auer, N.~Merkle, C.~Henry, and M.~Schmitt,
  ``{SAR-to-Optical Image Translation Based on Conditional Generative
  Adversarial Networks—Optimization, Opportunities and Limits},''
  \emph{Remote Sensing}, 2019.

\bibitem{zuo2021sar}
Z.~Zuo and Y.~Li, ``{A SAR-to-Optical Image Translation Method Based on
  PIX2PIX},'' in \emph{{IEEE International Geoscience and Remote Sensing
  Symposium IGARSS}}, 2021.

\bibitem{hwang_tripleGAN}
J.~Hwang and Y.~Shin, ``{Image Data Augmentation for SAR Automatic Target
  Recognition Using TripleGAN},'' in \emph{{2021 International Conference on
  Information and Communication Technology Convergence (ICTC)}}, 2021.

\bibitem{cui_dataAugment}
Z.~Cui, M.~Zhang, Z.~Cao, and C.~Cao, ``{Image Data Augmentation for SAR Sensor
  via Generative Adversarial Nets},'' 2019.

\bibitem{camus_refining}
B.~Camus, E.~Monteux, and M.~Vermet, ``{Refining Simulated SAR Images with
  Conditional GAN to Train ATR algorithms},'' in \emph{Actes de la
  conf{\'e}rence CAID}, 2020.

\bibitem{saxena2021generative}
D.~Saxena and J.~Cao, ``{Generative Adversarial Networks (GANs) Challenges,
  Solutions, and Future Directions},'' \emph{{ACM Computing Surveys (CSUR)}},
  2021.

\bibitem{diffusion_original}
J.~Sohl-Dickstein, E.~Weiss, N.~Maheswaranathan, and S.~Ganguli, ``{Deep
  Unsupervised Learning using Nonequilibrium Thermodynamics},'' in
  \emph{{International Conference on Machine Learning (PMLR)}}, 2015.

\bibitem{perera2023sar}
M.~V. Perera, N.~G. Nair, W.~G.~C. Bandara, and V.~M. Patel, ``{SAR Despeckling
  using a Denoising Diffusion Probabilistic Model},'' \emph{{IEEE Geoscience
  and Remote Sensing Letters}}, 2023.

\bibitem{ddpm}
J.~Ho, A.~Jain, and P.~Abbeel, ``{Denoising Diffusion Probabilistic Models},''
  \emph{Advances in Neural Information Processing Systems (NeurIPS)}, 2020.

\bibitem{mstar}
E.~R. Keydel, S.~W. Lee, and J.~T. Moore, ``{MSTAR Extended Operating
  Conditions: A Tutorial}.''\hskip 1em plus 0.5em minus 0.4em\relax
  International Society for Optics and Photonics, 1996.

\bibitem{inception_score}
T.~Salimans, I.~Goodfellow, W.~Zaremba, V.~Cheung, A.~Radford, and X.~Chen,
  ``{Improved Techniques for Training GANs},'' \emph{{Advances in Neural
  Information Processing Systems (NeurIPS)}}, 2016.

\bibitem{fid}
M.~Heusel, H.~Ramsauer, T.~Unterthiner, B.~Nessler, and S.~Hochreiter, ``{Gans
  Trained by a Two Time-Scale Update Rule Converge to a Local Nash
  Equilibrium},'' \emph{Advances in neural information processing systems},
  2017.

\bibitem{kid}
M.~Bi{\'n}kowski, D.~J. Sutherland, M.~Arbel, and A.~Gretton, ``{Demystifying
  MMD GANs},'' \emph{arXiv preprint arXiv:1801.01401}, 2018.

\bibitem{weng2021diffusion}
L.~Weng, ``What are diffusion models?'' \emph{lilianweng.github.io}, 2021.

\bibitem{nichol2021improved}
A.~Q. Nichol and P.~Dhariwal, ``{Improved Denoising Diffusion Probabilistic
  Models},'' in \emph{International Conference on Machine Learning}, 2021.

\bibitem{unet}
O.~Ronneberger, P.~Fischer, and T.~Brox, ``{U-Net: Convolutional Networks for
  Biomedical Image Segmentation},'' in \emph{{Medical Image Computing and
  Computer-Assisted Intervention--MICCAI}}, 2015.

\bibitem{group_normalization}
Y.~Wu and K.~He, ``{Group Normalization},'' in \emph{{Proceedings of the
  European conference on computer vision (ECCV)}}, 2018.

\bibitem{swish}
P.~Ramachandran, B.~Zoph, and Q.~V. Le, ``{Searching for Activation
  Functions},'' \emph{arXiv preprint arXiv:1710.05941}, 2017.

\bibitem{dcgan}
A.~Radford, L.~Metz, and S.~Chintala, ``{Unsupervised Representation Learning
  with Deep Convolutional Generative Adversarial Networks},'' \emph{arXiv
  preprint arXiv:1511.06434}, 2015.

\bibitem{sagan}
H.~Zhang, I.~Goodfellow, D.~Metaxas, and A.~Odena, ``{Self-Attention Generative
  Adversarial Networks},'' in \emph{{International Conference on Machine
  Learning}}.\hskip 1em plus 0.5em minus 0.4em\relax PMLR, 2019.

\bibitem{wgan}
M.~Arjovsky, S.~Chintala, and L.~Bottou, ``{Wasserstein Generative Adversarial
  Networks},'' in \emph{{International conference on machine learning}}.\hskip
  1em plus 0.5em minus 0.4em\relax PMLR, 2017.

\bibitem{spectralnorm}
T.~Miyato, T.~Kataoka, M.~Koyama, and Y.~Yoshida, ``{Spectral Normalization for
  Generative Adversarial Networks},'' \emph{{International Conference on
  Learning Representations (ICLR)}}, 2018.

\bibitem{inception_net}
C.~Szegedy, W.~Liu, Y.~Jia, P.~Sermanet, S.~Reed, D.~Anguelov, D.~Erhan,
  V.~Vanhoucke, and A.~Rabinovich, ``{Going Deeper with Convolutions},'' in
  \emph{{Proceedings of the IEEE Conference on Computer Vision and Pattern
  Recognition (CVPR)}}, 2015.

\end{thebibliography}

\end{document}